\documentclass[10pt,twocolumn,letterpaper]{article}

\usepackage{cvpr}
\usepackage{times}
\usepackage{epsfig}
\usepackage{graphicx}
\usepackage{amsmath}
\usepackage{amssymb}
\usepackage{booktabs}
\usepackage{bm}
\usepackage{caption}
\usepackage{subcaption}
\usepackage[breaklinks=true,bookmarks=false]{hyperref}

\cvprfinalcopy

\begin{document}

\title{DeepID3: Face Recognition with Very Deep Neural Networks}

\author{Yi Sun$^{1}$ ~~~~ Ding Liang$^{2}$ ~~~~  Xiaogang Wang$^{3,4}$  ~~~~ Xiaoou Tang$^{1,4}$\\
$^1$Department of Information Engineering, The Chinese University of Hong Kong\\
$^2$SenseTime Group\\
$^3$Department of Electronic Engineering, The Chinese University of Hong Kong\\
$^4$Shenzhen Institutes of Advanced Technology, Chinese Academy of Sciences\\
{\tt\small sy011@ie.cuhk.edu.hk} ~~  {\tt\small liangding@sensetime.com} \\
 ~~ {\tt\small xgwang@ee.cuhk.edu.hk} ~~  {\tt\small xtang@ie.cuhk.edu.hk}
}

\maketitle

\begin{abstract}
   The state-of-the-art of face recognition has been significantly advanced by the emergence of deep learning. Very deep neural networks recently achieved great success on general object recognition because of their superb learning capacity. This motivates us to investigate their effectiveness on face recognition.
   This paper proposes two very deep neural network architectures, referred to as DeepID3, for face recognition. These two architectures are rebuilt from stacked convolution and inception layers proposed in VGG net \cite{simonyan2014} and GoogLeNet \cite{szegedy2014} to make them suitable to face recognition. Joint face identification-verification supervisory signals are added to both intermediate and final feature extraction layers during training. An ensemble of the proposed two architectures achieves $99.53\%$ LFW face verification accuracy and $96.0\%$ LFW rank-1 face identification accuracy, respectively. A further discussion of LFW face verification result is given in the end.
\end{abstract}

\section{Introduction}

Using deep neural networks to learn effective feature representations has become popular in face recognition \cite{sun2013b,zhu2013,taigman2014a,zhu2014a,sun2014a,sun2014b,taigman2014b,zhu2014b,yi2014,sun2014c}. With better deep network architectures and supervisory methods, face recognition accuracy has been boosted rapidly in recent years. In particular, a few noticeable face representation learning techniques are evolved recently.
An early effort of learning deep face representation in a supervised way was to employ  face verification as the supervisory signal \cite{sun2013b}, which required classifying a pair of training images as being the same person or not. It greatly reduced the intra-personal variations in the face representation.
Then learning discriminative deep face representation through large-scale face identity classification (face identification) was proposed by DeepID \cite{sun2014a} and DeepFace \cite{taigman2014a,taigman2014b}. By classifying training images into a large amount of identities, the last hidden layer of deep neural networks would form rich identity-related features. With this technique, deep learning got close to human performance for the first time on tightly cropped face images of the extensively evaluated LFW face verification dataset \cite{huang2007a}. However, the learned face representation could also contain significant intra-personal variations.
Motivated by both \cite{sun2013b} and \cite{sun2014a}, an approach of learning deep face representation by joint face identification-verification was proposed in DeepID2 \cite{sun2014b} and was further improved in DeepID2+ \cite{sun2014c}. Adding verification supervisory signals significantly reduced intra-personal variations, leading to another significant improvement on face recognition performance. Human face verification accuracy on the entire face images of LFW was surpassed finally \cite{sun2014b,sun2014c}. Both GoogLeNet \cite{szegedy2014} and VGG \cite{simonyan2014} ranked in the top in general image classification in ILSVRC 2014. This motivates us to investigate whether the superb learning capacity brought by very deep net structures can also benefit face recognition.

Although supervised by advanced supervisory signals, the network architectures of DeepID2 and DeepID2+ are much shallower compared to recently proposed high-performance deep neural networks in general object recognition such as VGG and GoogLeNet. VGG net stacked multiple convolutional layers together to form complex features. GoogLeNet is more advanced by incorporating multi-scale convolutions and pooling into a single feature extraction layer coined inception \cite{szegedy2014}. To learn efficiently, it also introduced 1x1 convolutions for feature dimension reduction.

In this paper, we propose two deep neural network architectures, referred to as DeepID3, which are significantly deeper than the previous state-of-the-art DeepID2+ architecture for face recognition. DeepID3 networks are rebuilt from basic elements (\ie, stacked convolution or inception layers) of VGG net \cite{simonyan2014} and GoogLeNet \cite{szegedy2014}. During training, joint face identification-verification supervisory signals \cite{sun2014b} are added to the final feature extraction layer as well as a few intermediate layers of each network. In addition, to learn a richer pool of facial features, weights in higher layers of some of DeepID3 networks are unshared. Being trained on the same dataset as DeepID2+, DeepID3 improves the face verification accuracy from $99.47\%$ to $99.53\%$ and rank-1 face identification accuracy from $95.0\%$ to $96.0\%$ on LFW, compared with DeepID2+. The "true" face verification accuracy when wrongly labeled face pairs are corrected and a few hard test samples will be further discussed in the end.

\section{DeepID3 net}

\begin{figure}[t]
\begin{center}
\includegraphics[width=0.8\linewidth]{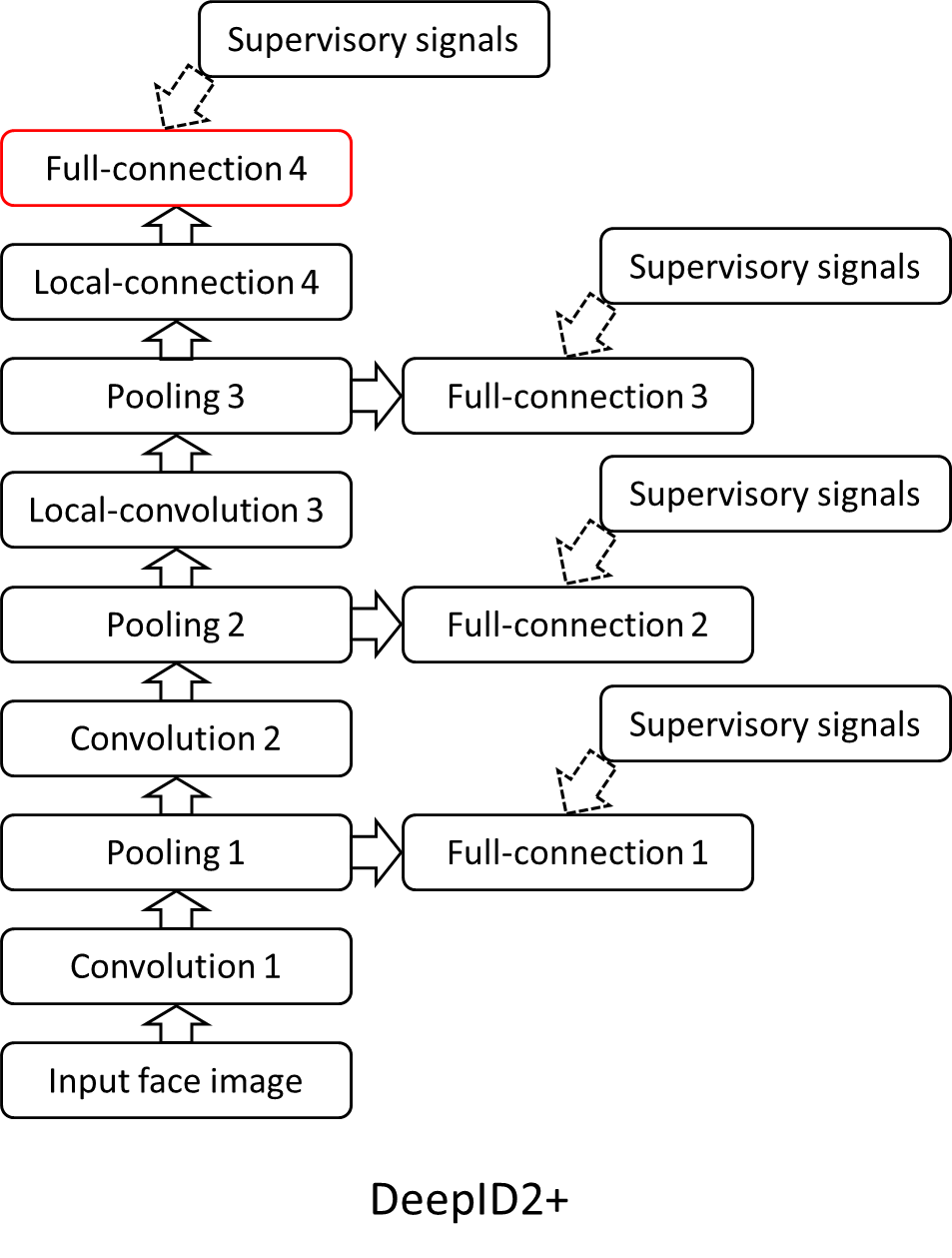}
\end{center}
\vspace{-0.15in}
\caption{Architecture of DeepID2+ net \cite{sun2014c}. Solid arrows show forward-propagation directions. Dashed arrows point the layers on which joint face identification-verification supervisory signals are added. The final feature extraction layer in red box is used for face recognition.}
\label{fig:struct_id2p}
\end{figure}

For the comparison purpose, we briefly review the previously proposed DeepID2+ net architecture \cite{sun2014c}. As illustrated in Fig. \ref{fig:struct_id2p}, DeepID2+ net has three convolutional layers followed by max-pooling (neurons in the third convolutional layer share weights in only local regions), followed by one locally-connected layer and one fully-connected layer. Joint identification-verification supervisory signals \cite{sun2014b} are added to the last fully-connected layer (from which the final features are extracted for face recognition) as well as a few fully connected layers branched out from intermediate pooling layers to better supervise early feature extraction processes.

\begin{figure}[t]
\begin{center}
\includegraphics[width=0.8\linewidth]{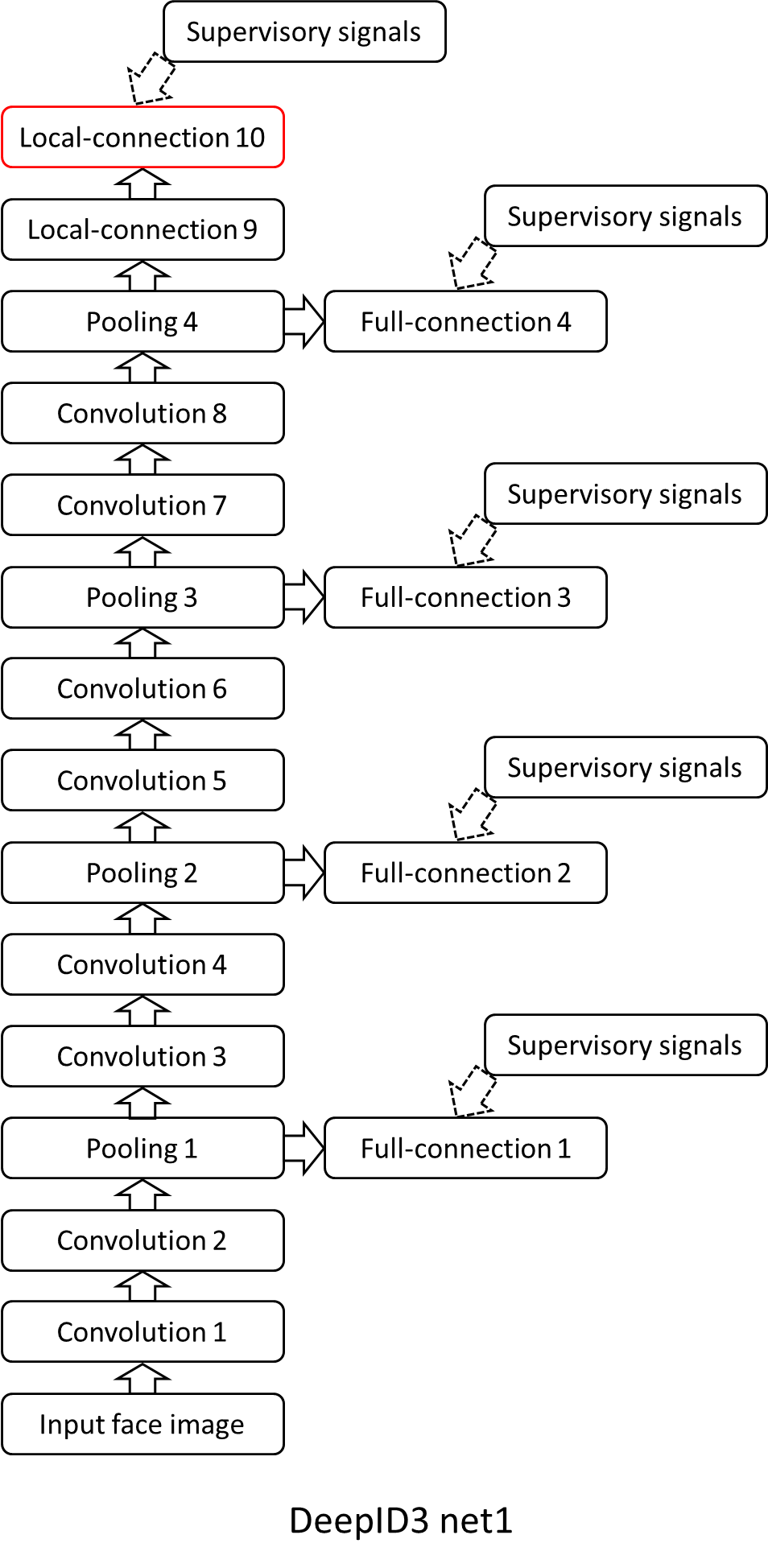}
\end{center}
\vspace{-0.15in}
\caption{Architecture of DeepID3 net1. Figure description is the same as Fig. \ref{fig:struct_id2p}.}
\label{fig:struct_id3_1}
\end{figure}

\begin{figure}[t]
\begin{center}
\includegraphics[width=0.8\linewidth]{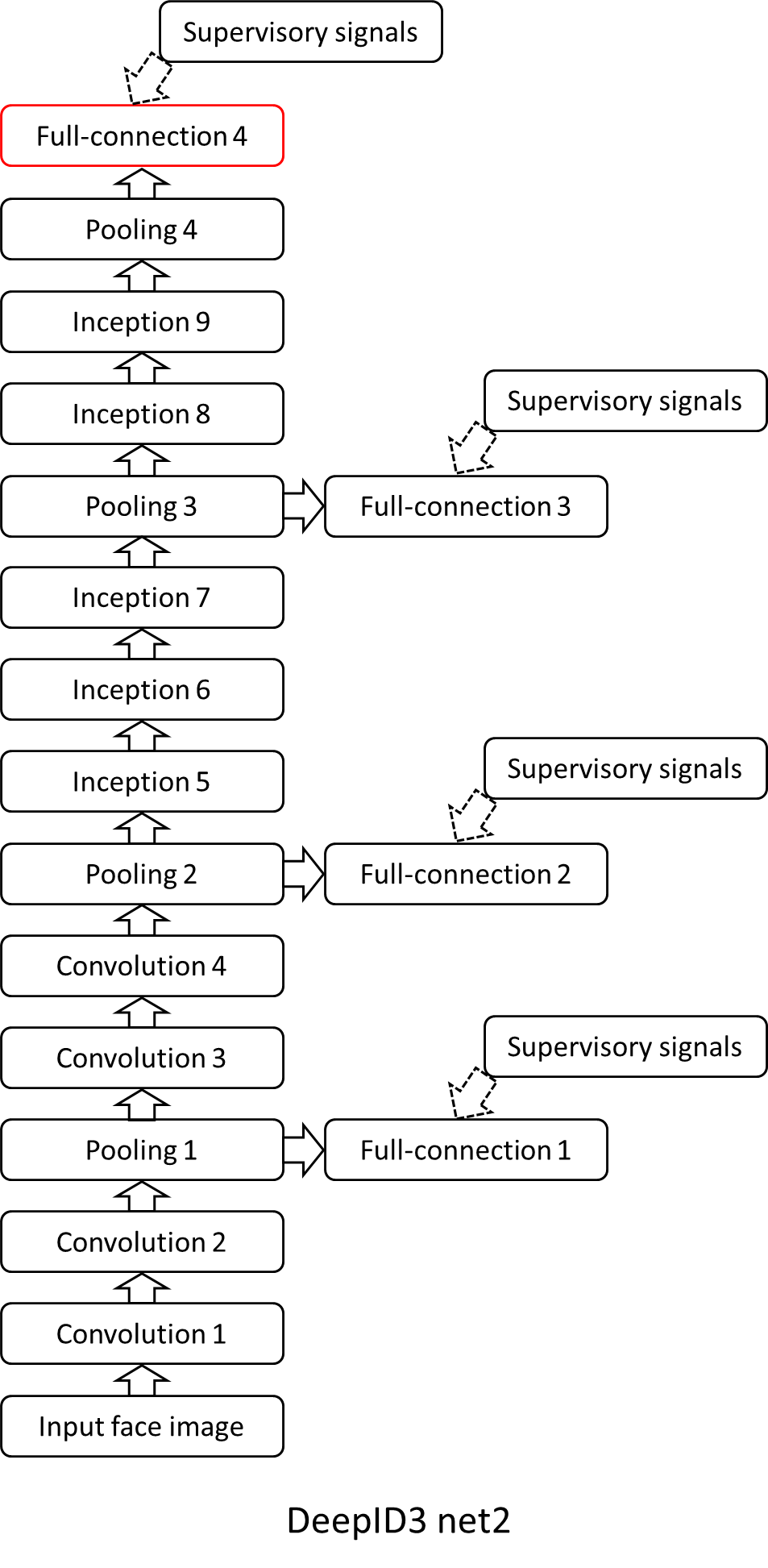}
\end{center}
\vspace{-0.15in}
\caption{Architecture of DeepID3 net2. Figure description is the same as Fig. \ref{fig:struct_id2p}.}
\label{fig:struct_id3_2}
\end{figure}

The proposed DeepID3 net inherits a few characteristics of the DeepID2+ net, including unshared neural weights in the last few feature extraction layers and the way of adding supervisory signals to early layers. However, the DeepID3 net is significantly deeper, with ten to fifteen non-linear feature extraction layers, compared to five in DeepID2+. In particular, we propose two DeepID3 net architectures, referred to as DeepID3 net1 and DeepID3 net2, as illustrated in Fig. \ref{fig:struct_id3_1} and Fig. \ref{fig:struct_id3_2}, respectively. The depth of DeepID3 net is due to stacking multiple convolution/inception layers before each pooling layer. Continuous convolution/inception helps to form features with larger receptive fields and more complex nonlinearity while restricting the number of parameters \cite{simonyan2014}.

The proposed DeepID3 net1 takes two continuous convolutional layers before each pooling layer. Compared to the VGG net proposed in previous literature \cite{simonyan2014,yi2014}, we add additional supervisory signals in a number of full-connection layers branched out from intermediate layers, which helps to learn better mid-level features and makes optimization of a very deep neural network easier. The top two convolutional layers are replaced by locally connected layers. With unshared parameters, top layers could form more expressive features with a reduced feature dimension. The last locally connected layer of our DeepID3 net1 is used to extract the final features without an additional fully connected layer.

DeepID3 net2 starts with every two continuous convolutional layers followed by one pooling layer as does in DeepID3 net1, while taking inception layers \cite{szegedy2014} in later feature extraction stages: there are three continuous inception layers before the third pooling layer and two inception layers before the fourth pooling layer. Joint identification-verification supervisory signals are added on fully connected layers following each pooling layer.

In the proposed two network architectures, rectified linear non-linearity \cite{nair2010} is used for all except pooling layers, and dropout learning \cite{hinton2012} is added on the final feature extraction layer. Although with significant depth, our DeepID3 networks are much smaller than VGG net or GoogLeNet proposed in general object recognition due to a restricted number of feature maps in each layer.

\begin{figure}[t]
\begin{center}
\includegraphics[width=\linewidth]{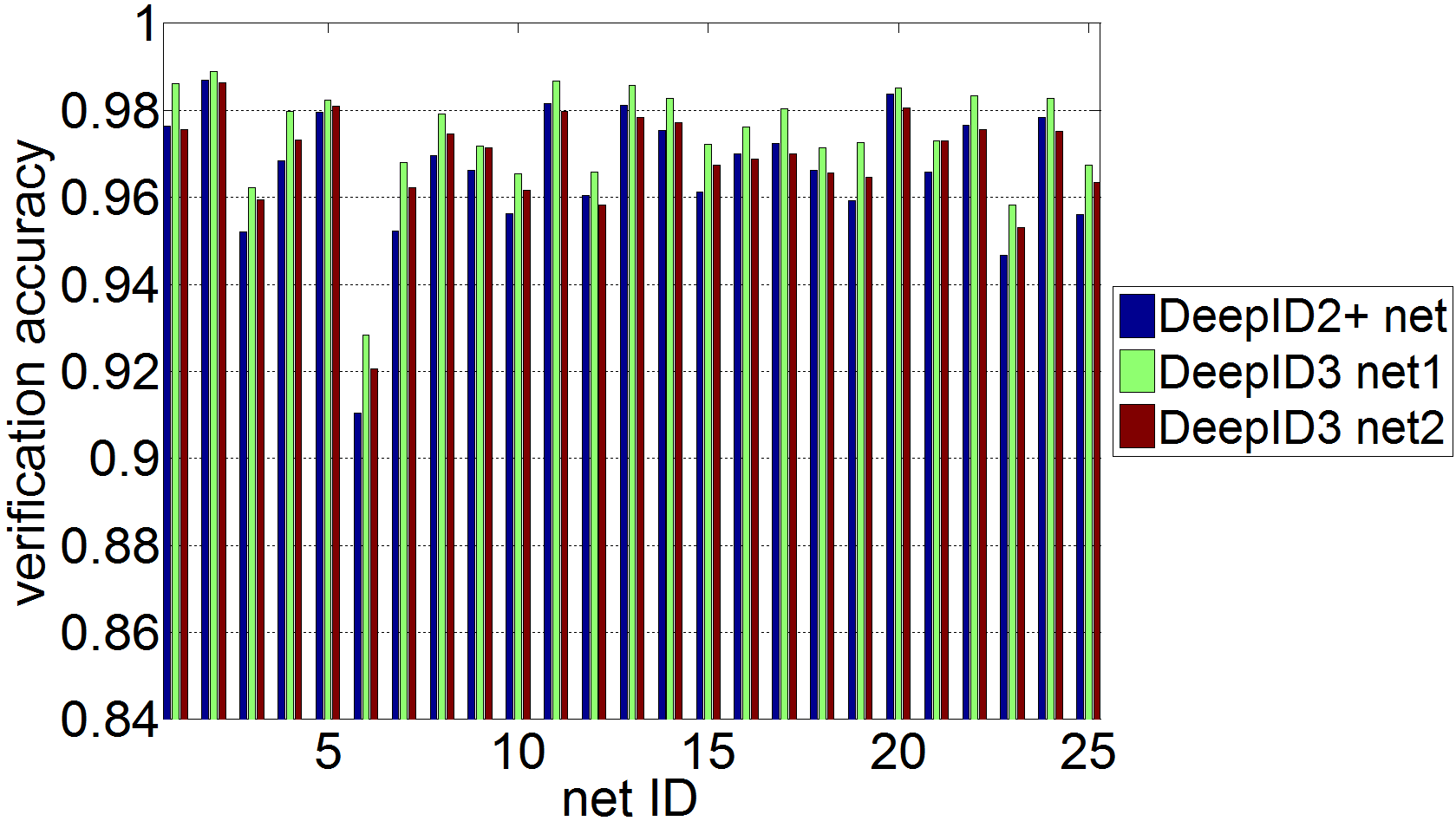}
\end{center}
\vspace{-0.15in}
\caption{LFW face verification accuracy of individual DeepID2+ and DeepID3 net trained on the same face regions in \cite{sun2014c}.}
\label{fig:net25}
\end{figure}

The proposed DeepID3 nets are trained on the same $25$ face regions as DeepID2+ nets \cite{sun2014c}, with each network taking a particular face region as input. These face regions are selected by feature selection in the previous work \cite{sun2014b}, which differ in positions, scales, and color channels such that different networks could learn complementary information. After training, these networks are used to extract features from respective face regions. Then an additional Joint Bayesian model \cite{chen2012} is learned on these features for face verification or identification. All the DeepID3 networks and Joint Bayesian models are learned on the same approximately $300$ thousand training samples as used in DeepID2+ \cite{sun2014c}, which is a combination of CelebFaces+ \cite{sun2014a} and WDRef \cite{chen2012} datasets, and tested on LFW \cite{huang2007a}. People in these two training data sets and the LFW test set are mutually exclusive. The face verification performance on LFW of individual DeepID3 net is compared to DeepID2+ net in Fig. \ref{fig:net25} on the $25$ face regions (with horizontal flipping), respectively. On average, DeepID3 net1 and DeepID3 net2 reduce the error rate by $0.81\%$ and $0.26\%$ compared to DeepID2+ net, respectively.

\section{Experiments}

To reduce redundancy, DeepID3 net1 and net2 are used to extract features on either the original or the horizontally flipped face region but not both. In test, feature extraction takes $50$ times of forward propagation with half from DeepID3 net1 and the other half from net2. These features are concatenated into a long feature vector of approximately $30,000$ dimensions. With PCA, it is reduced to $300$ dimensions on which a Joint Bayesian model is learned for face recognition.

\begin{figure}[!h]
\begin{center}
\includegraphics[width = 0.95\linewidth]{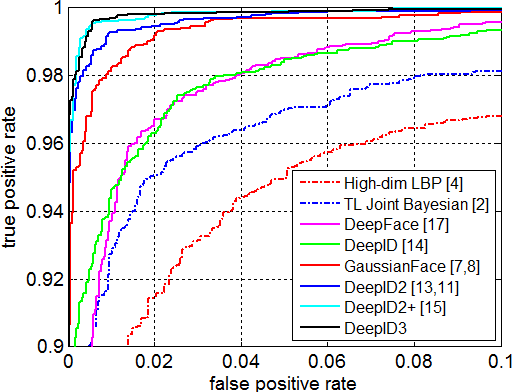}
\end{center}
\caption{ROC  of face verification on LFW.}
\label{fig:lfw}
\end{figure}

We evaluate DeepID3 networks under the LFW face verification \cite{huang2007a} and LFW face identification \cite{best-rowden2014,taigman2014b} protocols, respectively. For face verification, $6000$ given face pairs are verified to tell if they are from the same person. We achieve a mean accuracy of $\bm{99.53\%}$ under this protocol. Comparisons with previous works on mean accuracy and ROC curves are shown in Tab. \ref{tab:lfw} and Fig. \ref{fig:lfw}, respectively.

For face identification, we take one closed-set and one open-set identification protocols. For closed-set identification, the gallery set contains $4249$ subjects with a single face image per subject, and the probe set contains $3143$ face images from the same set of subjects in the gallery. For open-set identification, the gallery set contains $596$ subjects with a single face image per subject, and the probe set contains $596$ genuine probes and $9494$ imposter ones. Table \ref{tab:lfw_id} compares Rank-$1$ identification accuracy of closed-set identification and Rank-$1$ Detection and Identification rate (DIR) at a $1\%$ False Alarm Rate (FAR) of open-set identification, respectively. We achieve $\bm{96.0\%}$ closed-set and $\bm{81.4\%}$ open-set face identification accuracies, respectively.

\begin{table}[t]
\caption{Face verification on LFW.}
\label{tab:lfw}
\begin{center}
\begin{tabular}{p{100pt}|p{100pt}}
\toprule
method & accuracy (\%) \\
\midrule
High-dim LBP \cite{chen2013} & $95.17\pm1.13$ \\
TL Joint Bayesian \cite{cao2013} & $96.33\pm1.08$ \\
DeepFace \cite{taigman2014a} & $97.35\pm0.25$ \\
DeepID \cite{sun2014a} & $97.45\pm0.26$ \\
GaussianFace \cite{lu2014,lu2015} & $98.52\pm0.66$ \\
DeepID2 \cite{sun2014b,sun2014b2} & $99.15\pm0.13$ \\
DeepID2+ \cite{sun2014c} & $99.47\pm0.12$ \\
DeepID3 & \bm{$99.53 \pm 0.10$} \\
\bottomrule
\end{tabular}
\end{center}
\end{table}

\begin{table}[t]
\caption{Closed- and open-set identification tasks on LFW. }
\label{tab:lfw_id}
\begin{center}
\begin{tabular}{p{80pt}|p{50pt}|p{50pt}}
\toprule
method & Rank-1 (\%) & DIR $@$ $1\%$ FAR ($\%$) \\
\midrule
%BLS \cite{chen2013}* & $18.1$ & $7.89$ \\
COTS-s1 \cite{best-rowden2014} & $56.7$ & $25$ \\
COTS-s1+s4 \cite{best-rowden2014} & $66.5$ & $35$ \\
DeepFace \cite{taigman2014a} & $64.9$ & $44.5$ \\
WST Fusion \cite{taigman2014b} & $82.5$ & $61.9$ \\
DeepID2+ \cite{sun2014c} & $95.0$ & $80.7$ \\
DeepID3 & $\bm{96.0}$ & $\bm{81.4}$ \\
\bottomrule
\end{tabular}
\end{center}
\vspace{-0.1in}
\end{table}

\section{Discussion}

There are three test face pairs which are labeled as the same person but are actually different people as announced on the LFW website. Among these three pairs, two are classified as the same person while the other one is classified as different people by our DeepID3 algorithm. Therefore, when the label of these three face pairs are corrected, the actual face verification accuracy of DeepID3 is $99.52\%$. For DeepID2+ \cite{sun2014c}, its face verification accuracy before correcting the three wrong labels is $99.47\%$. However, DeepID2+ classified all the three wrongly labeled positive face pairs as different people. When these three wrong labels are corrected, the true face verification accuracy of DeepID2+ is also $99.52\%$ \cite{sun2014c}. DeepID3, although taking similar very deep architectures as VGG and GoogLeNet, does not improve over DeepID2+, with significantly shallower architecture, on the LFW face verification task. Whether those very deep architectures would take advantage of more training face data and finally surpass shallower architectures like DeepID2+ remains an open question.

\begin{figure}[!h]
\begin{center}
\includegraphics[width = 0.95\linewidth]{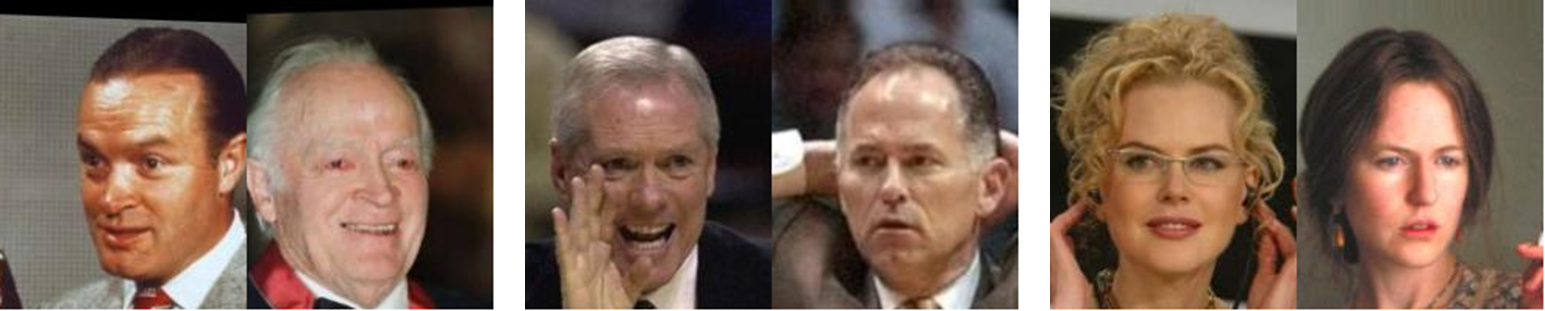}
\end{center}
\caption{Common false negatives in DeepID series algorithms.}
\label{fig:fn}
\end{figure}

\begin{figure}[!h]
\begin{center}
\includegraphics[width = 0.95\linewidth]{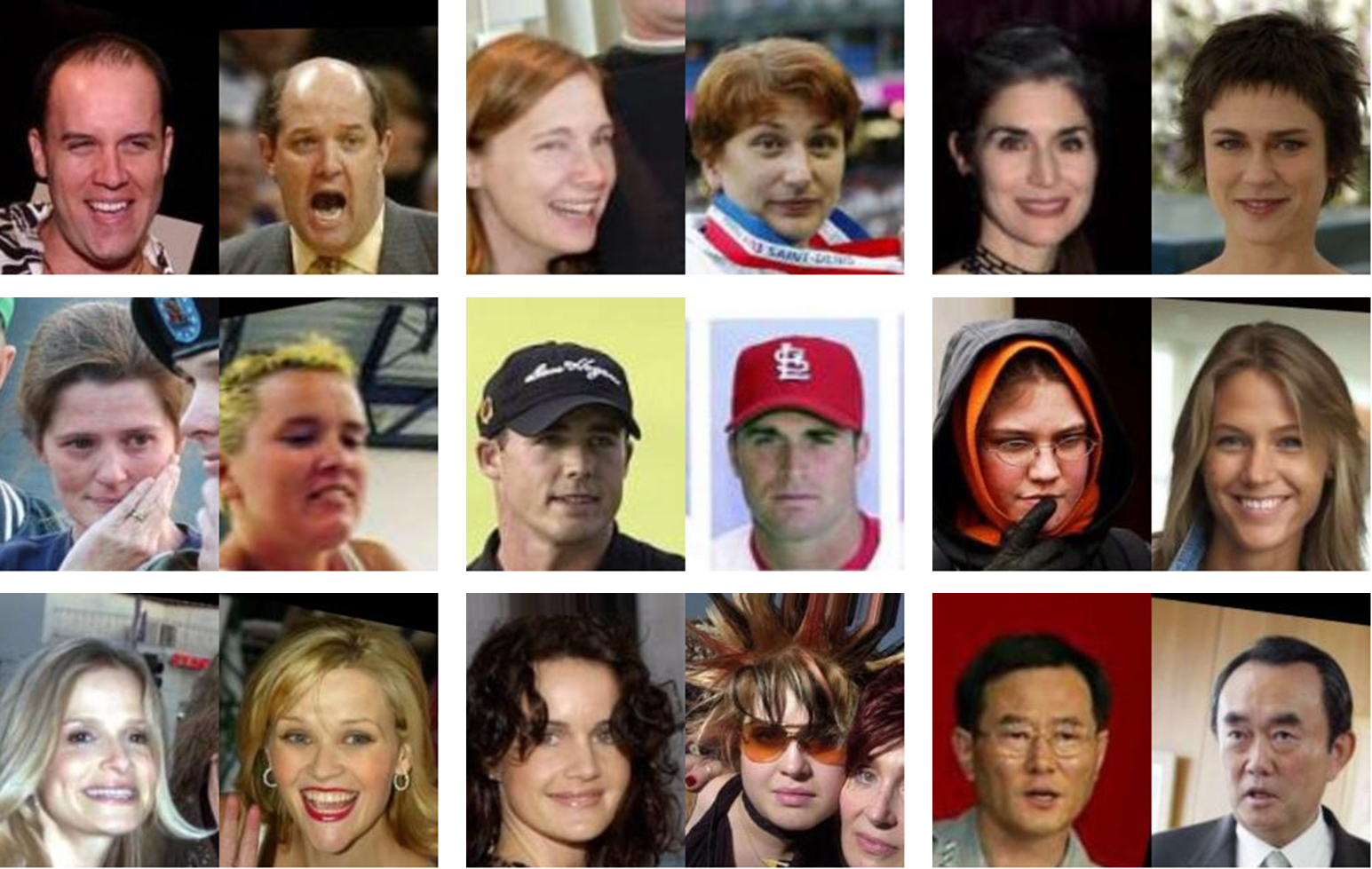}
\end{center}
\caption{Common false positives in DeepID series algorithms.}
\label{fig:fp}
\end{figure}

We examine the test face pairs in LFW which are wrongly classified by all the DeepID series algorithms including DeepID \cite{sun2014a}, DeepID2 \cite{sun2014b,sun2014b2},DeepID2+ \cite{sun2014c}, and DeepID3. There are nine common false positives and three common false negatives in total, around half of all wrongly classified face pairs by DeepID3. The three face pairs labeled as the same person but being classified as different people are shown in Fig. \ref{fig:fn}. The first pair of faces show great contrast of ages. The second pair is actually different people due to errors in labeling. The third one is an actress with significantly different makeups. Fig. \ref{fig:fp} shows the nine face pairs labeled as different people while being classified as the same person by algorithms. Most of them look similar or have interference such as occlusions.

\section{Conclusion}

This paper proposes two significantly deeper neural network architectures, coined DeepID3, for face recognition. The proposed DeepID3 networks achieve the state-of-the-art performance on both LFW face verification and identification tasks. However, when a few wrong labels in LFW are corrected, the improvement of DeepID3 over DeepID2+ on LFW face verification vanished. The effectiveness of very deep neural networks would be further investigated on larger scale training data in the future.

{\small
\bibliographystyle{ieee}
\bibliography{deepid3}
}

\end{document}